\documentclass[runningheads]{llncs}
\usepackage{graphicx}
\usepackage{wrapfig}%
\usepackage{amsmath, amssymb}
\usepackage{bm}
\usepackage{subfigure}

\DeclareMathOperator{\round}{round}
\DeclareMathOperator{\BatchNorm}{BatchNorm}
\DeclareMathOperator{\Sign}{Sign}

\begin{document}

\title{B-DCGAN: Evaluation of Binarized DCGAN for FPGA}

\author{Hideo Terada \inst{1} \and Hayaru Shouno \inst{2}}
\institute{
  Open Stream, Inc., 2-7-1 Nishi-Shinjuku, Shinjuku, Tokyo 163-0709, Japan \\
  \email{terada.h@opst.co.jp}
  \and
  Graduate School of Informatics and Engineering, University of Electro-Communications, Chofugaoka 1-5-1, Chofu, Tokyo 182-8585, Japan \\
  \email{shouno@uec.ac.jp}
}

\authorrunning{H. Terada et al.}

\maketitle

\begin{abstract}
In this work, we demonstrate an implementation of Deep Convolution Generative Adversarial Network (DCGAN), into a Field Programmable Gate Array (FPGA).
In order to implement the DCGAN, we modified the DCGAN model with binary weights and activations, and with using integer-valued operations in the forwarding path(train-time and run-time). We call the modified one as Binary-DCGAN(B-DCGAN)
Using the B-DCGAN, we do a feasibility study of FPGA's characteristics and performance for Deep Learning.
Because the binarization and using integer-valued operation reduce the memory capacity and the number of the circuit gates, it is very effective for FPGA implementation. On the other hand, these reductions in the model might decrease the quality of the generated data. So we investigate the influence of these reductions.
\end{abstract}

\section{Introduction}
We try to implement a deep neural network in the edge computing environment for real-world applications such as the IoT(Internet of Things), the FinTech, to utilize the significant achievement of Deep Learning in recent years. Especially, we now focus on algorithm implementation on FPGA, because it is one of the promising devices for low-cost and low-power implementation in edge computing.

A well-known fact, Graphics Processing Unit(GPU) is now the most common computing resources for Deep Neural Networks(DNNs).
GPU is one of the most useful devices to execute massive scale numerical operations such as DNNs.
Furthermore, thanks to the effort of world-wide researchers and developers, now there are many software tools, libraries, and frameworks suitable for GPU, so it is easy to write the DNNs software for GPU.
However, GPU has very high power consumption, so that it needs an abundant power supply and cooling equipment. Thus, it is hard to apply GPUs in the small edge machines for IoT.

As a way for low-power and low-cost dedicated computation, Field Programmable Gate Array(FPGA) is gathering attention recently again, since its cost performance has improved and the prices of related software tools have also declined. 

While FPGA is user-programmable hardware, its development in the early times was pretty tricky and complicated, because the program requires `low-level' notations such as Hardware Description Language(HDL) or schematic circuit diagrams, so the developers had to have these kinds of special skills.

However, nowadays, owing to the High-Level Synthesis(HLS) technique has been evolving, the developer programs the FPGA using a high-level programming language such as C/C++. It is a much easier way for the developer who is not an expert of hardware.

\subsection{The Problem of Using FPGA for DNN}
The algorithm development using FPGA has the following problems:
First, the number of gates and the capacity of fast memory is much smaller than that of the GPU. While it is available to use either multi FPGAs, or FPGA with external memories, however, the processing speed will drop considerably because of the communication overheads.
Second,  the more the algorithm uses multiplication, division, or floating-point value of those operations, the much larger gates it will consume.
Third, if the circuit is not optimized appropriately in the hardware-specific way, such as parallelizing or pipelining, the FPGA could get slower speed than a CPU with the same clock range.

As the solution for these problems, various researches make the circuit scale smaller for DNNs:
Courbariaux {\it{et al.}} showed that image classification using the neural networks with binary weights and activations(BNN) is enough possible~\cite{2016arXiv160202830C-BNN}.
Moreover, they showed the method to train the BNN.
Umuroglu {\it{et al.}} presented the FINN\cite{2016arXiv161207119U-FINN}, a framework for building fast and flexible FPGA accelerators that enable efficient mapping of binarized neural networks to hardware.
Adelouahab {\it{et al.}} published a good servey\cite{abdelouahab:hal-01695375-Survey-Accl-FPGA}, which reports various types of implementation techniques of CNN on FPGA.
Cheng {\it{et al.}} also published another good suvey\cite{2017arXiv171009282C-Survey-Model-Comp}, in which they discussed the recent techniques for compacting and accelerating CNNs in detail.


\subsection{Contributions}
This work makes the following contributions:
\begin{itemize}
\item{We introduce binarized \& integer-valued deep convolutional conditional generative adversarial networks. Hereafter we call it as B-DCGAN.}
\item We show the quality change of output from B-DCGAN Generator when the extent of binarization was changed.
\end{itemize}
\section{B-DCGAN: DCGAN With Binarized Generator}
Binary Deep Convolutional \& Conditional Generative Adversarial Networks(B-DCGAN) is a modified version of DCGAN \cite{2015arXiv151106434R-DCGAN}.
In the B-DCGAN, the Discriminator network is a vanilla network as the normal DCGAN; that is, it uses real-valued operation.
However, the Generator network adopts the binary weights, binary activations, and integer-valued operations. 

\subsection{Network Structure}
\begin{figure}
    \begin{center}%
        \includegraphics[height=12cm, angle=90]{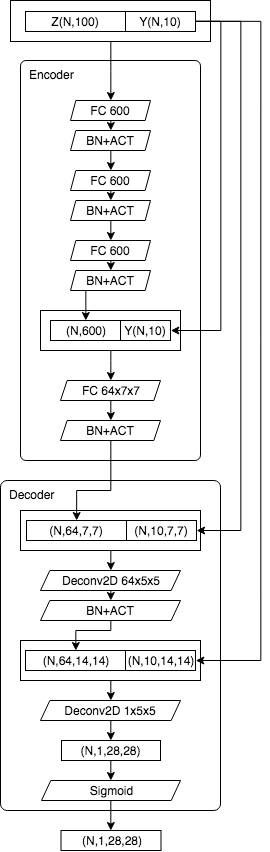}
        \caption{Scematic diagram of B-DCGAN Generator}
        \label{fig:gen}
    \end{center}%
\end{figure}

Fig.\ref{fig:gen} is a schematic diagram of the network structure of B-DCGAN Generator. The first half of the Generator is consist of full connection layers, which is called Encoder. The last half is named Decoder, is consist of deconvolution layers.

We show the details of these layers in the following sections.

\subsection{Integer-Valued Inputs}
\label{subsection:integer-valued-inputes}
The Generator's inputs are $\bm{z}$ and $\bm{y}$. In vanilla GAN, $\bm{z}$ is 1D-vector of random floating point values in range $[-1.0 \sim +1.0]$. In B-DCGAN, we use integer value $\bm{z^i}$ for input as follows:
\begin{alignat}{3}
    \bm{z} &= (z_1, z_2, \cdots,  z_n)&\\
    \bm{z^i} &= (z^i_1, z^i_2, \cdots, z^i_n)&\\
    z^i_k &= \round(A \cdot z_k)&(k=1,2,\cdots,n)\\
    A &=2^{h-1} - 1&(h=2,3,4,\cdots).
\end{alignat}
The B-DCGAN treats the conditional input method like as CGAN\cite{2014arXiv1411.1784M-CGAN}.
The $\bm{y}$ is one-hot vector representation of class label to be generated.
It is also converted into integer value $\bm{y^i}$:
\begin{align}
    \bm{y^i} = A\cdot\bm{y}=(y^i_1, y^i_2, \cdots, y^i_l, \cdots, y^i_c)
    \;\;(l=1,2,\cdots,c), 
\end{align}
where the constant $A$ is a hyper-parameter of B-DCGAN. $z^i_k$ and $y^i_l$ is integer values represented by signed $k$ bits. The integer value is very advantageous from a hardware perspective because the number of circuit logics of integer process is much smaller than those of floating point in FPGA.
We have investigated the influence of the value $A$ setting for quality of Generator's output as we explain in Section \ref{sec:Evaluation}.

\subsection{Binarized Full Connection Layer(B-FC)}
\label{B-FC}
Implementation of Full Connection layer in B-DCGAN is based on those of BNN\cite{2016arXiv160202830C-BNN}.
During the backward pass(at train-time), its weight value is real-valued variable the same as the usual full connection.
During the forward path calculation(both at run-time and train-time), we treat the weight as binary, that is, the value is constrained to either +1 or -1 as follows:

\begin{equation}
\label{eq:bin-1}
    x^b = \Sign(x) = \begin{cases}
    +1, & \text{if $x>=0$} \\
    -1, & \text{otherwise}
    \end{cases}
\end{equation}

\noindent

We have investigated the influence of this binarization for quality of Generator's output as we explain in Section \ref{sec:Evaluation}.

\subsection{Binarized Batch Normalization + Activation Layer(B-BNA)}
\label{B-BNA}
In our B-DCGAN structure, the batch normalization layer the non-linear activation layer, so we treat these two layers as one layer calculation. This consideration is based on the `Batchnorm-activation as Threshold' in the FINN paper(Umuroglu et al. \cite{2016arXiv161207119U-FINN}); that is, we have modified it as integer-valued version.

Let $a_j$ is the output of $j$th neuron in the previous layer, and $\Theta_j=(\gamma_j, \mu_j, i_j, B_j)$ is the batch normalization parameter set learned during the training phase. The output of this layer $a^b_j$ in computed as:
\begin{align}
    a^b_j = \begin{cases}
    +1,& \text{if $a_j >= \tau^b_j$}\\
    -1,& \text{otherwise}
      \end{cases}
\end{align}
The threshold $\tau^b_j$ is computed by solving $BatchNorm(\tau_j, \Theta_j)=0$ and rounding:
\begin{align}
    \BatchNorm(\tau_j, \Theta_j) &= \gamma_j\cdot(\tau_j-\mu_j)\cdot i_j + B_j=0\\
    \therefore 
    \tau_j &= \mu_j - (B_j/(\gamma_j\cdot i_j))\\
    \tau^b_j &= \round(\tau_j)
\end{align}

During the training phase, $\tau^b_j$ value is always modified according to the change of $BatchNorm$ parameters; however, in the run-time phase, the value can be treated as a fixed value.

\subsection{Binarized Deconvolution Layer(B-Deconv)}
The behavior of the binarized deconvolution layer(B-Deconv) is similar to those of the B-FC layer described at \ref{B-FC}.
That is, during the backward path calculation at the train-time, the weight values of filters are real-valued variable same as the normal deconvolution.
However, in the forward path calculation both at the run-time and train-time, we treat the weight values as binarized to either $+1$ or $-1$ according to the equation (\ref{eq:bin-1}).

We have also investigated the influence of using B-Deconv for output quality of the Generator(see \ref{sec:S3-1}).

\section{Evaluation}
\label{sec:Evaluation}
\subsection{Experimental Setup}
\subsubsection{Scenario}
To exam B-BDCGAN, we arranged several configurations of network binarization.
We present a configuration as a set of flags and hyper-parameters described in Table.\ref{table:setup-scenario}.
We call it `{\bf Scenario}.'
The `Input as integer' flag represents whether the input is integer-valued(Y) or real-valued(n).
Other flags correspond with whether each layer is binarized(Y) or not(n).
The '$A$ value' is input scale value that is only effective when the `Input as integer' is positive(Y), had explained in \ref{subsection:integer-valued-inputes}.

\begin{table}[h]
\centering
\caption{ Setup Scenario. 'Y' is positive, 'n' is negative.}
\label{table:setup-scenario}
\begin{tabular}{|l|c|c|c|c|c|c|c|}
\hline
 & \multicolumn{7}{c|}{\textbf{Scenario \#}} \\ \hline
                    & S0 & S1-1 & S1-2 & S2-1 & S2-2 & S3-1 & S3-2 \\ \hline
\multicolumn{8}{|l|}{\textbf{Encoder}}                          \\ \hline
Input as integer & n  & Y    & Y    & Y    & Y    & Y    & Y    \\ \hline
$A$ value        & -  & 1    & 1    & 127  & 4095 & 1    & 1    \\ \hline
B-FC               & n  & Y    & Y    & Y    & Y    & Y    & Y    \\ \hline
B-BNA-1          & n  & n    & Y    & Y    & Y    & Y    & Y    \\ \hline
\multicolumn{8}{|l|}{\textbf{Decoder}}                          \\ \hline
B-Deconv-1       & n  & n    & n    & n    & n    & Y    & Y    \\ \hline
B-BNA-2          & n  & n    & n    & n    & n    & Y    & Y    \\ \hline
B-Deconv-2       & n  & n    & n    & n    & n    & n    & Y    \\ \hline
\end{tabular}
\end{table}

\subsubsection{Constant of Hyper-Parameters}
We adopt the following constant as the hyper-parameters for the network:
The number of units in Full Connection layer is $600$.
The Kernel sizes in Deconvolution layer-1, and -2 is $5\times 5$.
The number of filters of Deconvolution layer-1 is $64$.
Also, the number of filters of Deconvolution layer-2 is $1$.

\begin{figure}[h]
    \begin{center}
    \subfigure[initial]{
        \includegraphics[width=0.2\columnwidth]{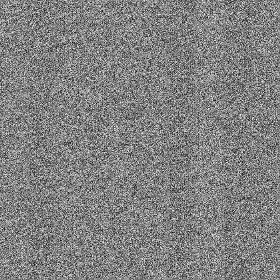}}
    \subfigure[halfway-1]{
        \includegraphics[width=0.2\columnwidth]{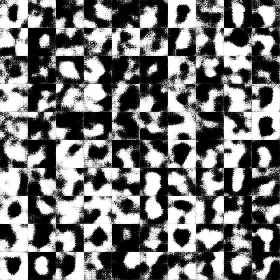}}
    \subfigure[halfway-2]{
        \includegraphics[width=0.2\columnwidth]{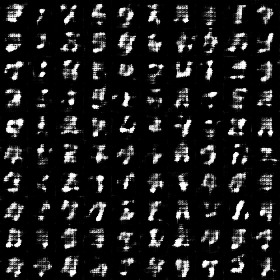}}
    \subfigure[peak]{
        \includegraphics[width=0.2\columnwidth]{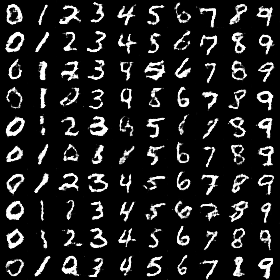}}
    \caption{Generator's output images in the middle of training\newline}
    \label{fig:image-middle-of-training}
    \end{center}
\end{figure}

\begin{figure}[ht]
    \begin{center}
        \includegraphics[width=4cm]{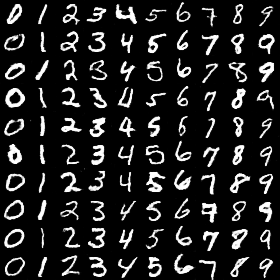}
        \caption{peak image of S0 (baseline)}
        \label{fig:S0-peak}
    \end{center}
\end{figure}

\begin{figure}[ht]
    \begin{minipage}{0.5\hsize}
        \centering
        \includegraphics[width=4cm]{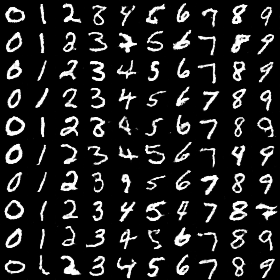}
        \caption{peak image of S1-1}
        \label{fig:S1-1-peak}
    \end{minipage}
    \begin{minipage}{0.5\hsize}
        \centering
          \includegraphics[width=4cm]{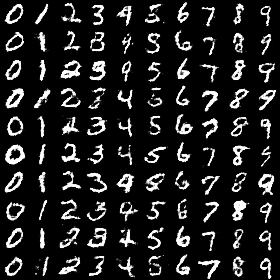}
        \caption{peak image of S1-2}
        \label{fig:S1-2-peak}
    \end{minipage}
\end{figure}

\begin{figure}[ht]
    \begin{minipage}{0.5\hsize}
        \centering
        \includegraphics[width=4cm]{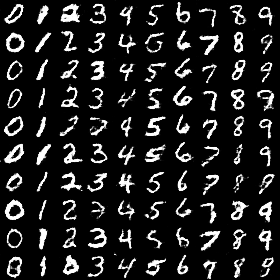}
        \caption{peak image of S2-1}
        \label{fig:S2-1-peak}
    \end{minipage}
    \begin{minipage}{0.5\hsize}
        \centering
          \includegraphics[width=4cm]{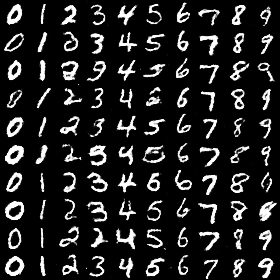}
        \caption{peak image of S2-2}
        \label{fig:S2-2-peak}
    \end{minipage}
\end{figure}

\subsubsection{Training}
We apply the MNIST\cite{Lecun98gradient-basedlearning} dataset for training. 
We make the training program by python using Theano\cite{2016arXiv160502688full-Theano} library and Lasagne \cite{lasagne} library, and partially using BinaryNet \cite{2016arXiv160202830C-BNN} code and DCGAN \cite{2015arXiv151106434R-DCGAN}  code. We run the program on Ubuntu Linux with NVIDIA Titan-X GPU.
Our program code is available on-line\footnote{{\tt https://github.com/hterada/b-dcgan}}.

We adopt a stochastic gradient descent(SGD) method as the training method, in which we set the number of mini-batches as $128$, and the initial learning rate as $0.0001$ that was linearly decreased down to 0 with the decay step of $0.0001/3000=3\times 10^-8$.

We judge the finish of training of each Scenario through visual assessment for the quality of output images from the Generator.
That is, at each training iteration, the program generates output image and current parameters from the Generator at that time to individual files, so we can examine these files in order to find when the quality reaches its peak.
For example, Fig.\ref{fig:image-middle-of-training} shows output images in the middle of training and the peak image. As this figure shows, the quality is rising according to the iterations of training.

\subsubsection{FPGA implementation}
After the training of the DCGAN network model, we extracted the set of model parameters of the peak quality.
These parameters were written in python pickle formatted files with filename extension `{\tt{.jl}}'.
We made a dedicated python program called `{\tt{model\_to\_ch.py}}', which converts `{\tt{.jl}}' file to C++ header files, where the model parameters are expressed as C++ constant variables.
The building of the B-DCGAN Generator in Xilinx Vivado HLS for FPGA requires these header files.


A binarized parameter represented by +1 or -1 in training model should be represented by 1 or 0(1-bit value) in FPGA, so the `{\tt{model\_to\_ch.py}}' is converted $\pm 1$ parameters into such bit-mapped expression.

\subsection{Training Results in Scenarios}

\subsubsection{S0: Not Binarized (Vanilla) DCGAN}

\label{sec:S0}
The Scenario S0 is the baseline of image quality.
Figure.\ref{fig:S0-peak} shows output image of peak quality(we call it '\textbf{peak image}') from the trained model configured in S0.

\subsubsection{S1-1 vs S1-2: Encoder only binarization}

\label{sec:S1}
The Scenario S1-1 and S1-2 are experiments to examine the influence of binarization of Encoder.
\begin{itemize}
    \item The S1-1 makes input as integer and binarize only FC layer(see \ref{B-FC}).
    \item The S1-2 makes input as integer and binarize FC and BNA layer(see \ref{B-BNA}).
\end{itemize}
\noindent
Figure.\ref{fig:S1-1-peak} shows peak of S1-1, and Figure.\ref{fig:S1-2-peak} shows peak of S1-2.

\subsubsection{S2-1 vs S2-2: A value change}

\label{sec:S2}
The Scenario S2-1 and the S2-2 are experiments to examine the influence of '$A$' value of Encoder.
\begin{itemize}
    \item In the S2-1, we set $A$ value as $127$.
    \item In the S2-2, we set $A$ value to $4095$.
\end{itemize}
Figure.\ref{fig:S2-1-peak} and Figure.\ref{fig:S2-2-peak} show the peak generations of
the S2-1 and S2-2, respectively.

\subsubsection{S3-1 vs S3-2: Encoder \& Decoder binarization}

\begin{figure}[ht]
  \centering
  \includegraphics[width=4cm]{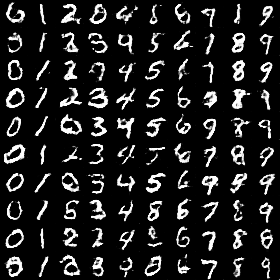}
  \caption{peak image of S3-1}
  \label{fig:S3-1-peak}
\end{figure}

\begin{figure}[ht]
  \begin{minipage}{0.325\hsize}
    \centering
    \includegraphics[width=4cm]{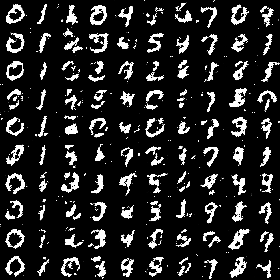}
  \end{minipage}
  \begin{minipage}{0.325\hsize}
    \centering
    \includegraphics[width=4cm]{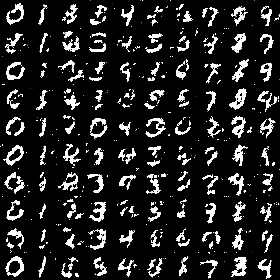}
  \end{minipage}
  \begin{minipage}{0.325\hsize}
    \centering
    \includegraphics[width=4cm]{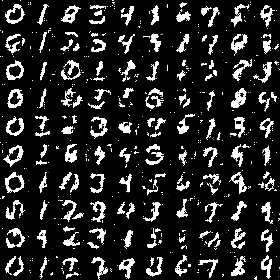}
  \end{minipage}
  \caption{peak candidates of S3-2}
  \label{fig:S3-2-cands}
\end{figure}

The Scenario S3-1 and S3-2 are experiments to examine influence of Decoder binarization.
\begin{itemize}
\item The S3-1 adopts the binarize Encoder, B-Deconv-1 and B-BNA-2 in Decoder \label{sec:S3-1}
\item The S3-2 adopts the binarize Encoder, B-Deconv-1, B-BNA-2 and B-Deconv-2 in Decoder; that is all layers in Decoder \label{sec:S3-2}
\end{itemize}
Figure.\ref{fig:S3-1-peak} shows the peak of the S3-1.
Figure.\ref{fig:S3-2-cands} shows the results of candidates for peak of the S3-2.
According to the training results, in the S3-2 Scenario, the quality of the output image looks very unstable.
So we could not decide a single peak in clear. Therefore, we picked up some candidates for the peak.

\section{Conclusion \& Discussion}
We have introduced B-DCGAN; that is DCGAN with binary weights and activations, and integer-valued input. 
We have conducted several scenarios of experiments on the Theano/Lasagne libraries,
which show that it is possible to binarize DCGAN model, 
moreover, show what extent is the binarization available with keeping output quality acceptable.

According to the results of these experiments, the last layer B-Deconv-2 mainly have the initiative for output image quality.
So the last layer has to operate in the real-valued process. Except for the last layer, other layers can be fully binarized.

According to the results in each Scenario above, we conduct the following speculations:
\begin{enumerate}
\item The output quality is affected very few by only binarization of the Encoder(cf. S0, S1-1, S1-2).
\item No influence for quality appears when the $A$ value has changed(cf. S2-1, S2-2).
\item The output quality is somewhat degraded by binarization both the Encoder and partially of the Decoder(cf. S3-1)
\item The output quality is degraded seriously by full binarization of both the Encoder and the Decoder(cf. S3-2)
\item The output quality chiefly depends on the Deconv-2 layer because of the quality difference between S3-1 and S3-2.
\end{enumerate}
At least on the MNIST dataset, we can select the S3-1 as the best Scenario for the B-DCGAN.


\section*{Acknowledgements}
We would like to thank the UEC Shouno lab\footnote{{\tt http://daemon.inf.uec.ac.jp/ja/}} members: Satoshi Suzuki and Aiga Suzuki for theoretical and technical discussion;
Seigo Kawamura, Kurosaka Mamoru, Yoshihiro Kusano, Toya Teramoto and Akihiro Endo for kind technical assistance and humor.
We thank Kazuhiko Yoshihara and all the members of Open Stream, Inc. 
We also thank the developers of Theano, Lasagne, and Python environment.
This work is supported under the funds of Open Stream, Inc.\footnote{{\tt https://www.opst.co.jp/}}

\bibliographystyle{splncs04}
\bibliography{bdcgan}

\end{document}